\documentclass[conference]{IEEEtran}
\IEEEoverridecommandlockouts

\usepackage{multirow}
\usepackage{hhline}
\usepackage{verbatim}
\usepackage{cite}
\usepackage{enumitem}
\usepackage{algorithm}
\usepackage{algpseudocode}
\usepackage{graphics} 
\usepackage{graphicx}
\usepackage{subfig}
\usepackage{url}
\usepackage{balance}
\usepackage{siunitx}
\usepackage{booktabs}
\usepackage{hyperref}
\usepackage[font={small}]{caption}
\usepackage{amsmath,amssymb,amsfonts}

\newcommand{\R}{\mathbb{R}}

\begin{document}

\title{EXCON: Extreme Instance-based Contrastive Representation Learning of Severely Imbalanced Multivariate Time Series for Solar Flare Prediction}
\author{\IEEEauthorblockN{Anonymous}}

\author{
    \IEEEauthorblockN{Onur Vural, Shah Muhammad Hamdi, Soukaina Filali Boubrahimi}
    \IEEEauthorblockA{Department of Computer Science, Utah State University, Logan, UT 84322, USA\\
    Emails: \{onur.vural, s.hamdi, soukaina.boubrahimi\}@usu.edu
}}
\maketitle

\begin{abstract}

In heliophysics research, predicting solar flares is crucial due to their potential to substantially impact both space-based systems and Earth's infrastructure. Magnetic field data from solar active regions, recorded by solar imaging observatories, are transformed into multivariate time series to enable solar flare prediction using temporal window-based analysis. In the realm of multivariate time series-driven solar flare prediction, addressing severe class imbalance with effective strategies for multivariate time series representation learning is key to developing robust predictive models. Traditional methods often struggle with overfitting to the majority class in prediction tasks where major solar flares are infrequent. This work presents EXCON, a contrastive representation learning framework designed to enhance classification performance amidst such imbalances. EXCON operates through four stages: (1) obtaining core features from multivariate time series data; (2) selecting distinctive contrastive representations for each class to maximize inter-class separation; (3) training a temporal feature embedding module with a custom extreme reconstruction loss to minimize intra-class variation; and (4) applying a classifier to the learned embeddings for robust classification. The proposed method leverages contrastive learning principles to map similar instances closer in the feature space while distancing dissimilar ones, a strategy not extensively explored in solar flare prediction tasks. This approach not only addresses class imbalance but also offers a versatile solution applicable to both univariate and multivariate time series across binary and multiclass classification problems. Experimental results, including evaluations on the benchmark solar flare dataset and multiple time series archive datasets with binary and multiclass labels, demonstrate EXCON's efficacy in enhancing classification performance and reducing overfitting.

\end{abstract}


\section{Introduction} \label{sec:intro}
In heliophysics research, solar flares are rapid, intense bursts of radiation from the Sun's surface, caused by the release of magnetic energy stored in the Sun's atmosphere. Flares are classified logarithmically according to their peak soft X-ray flux in the 1–8 Å wavelength band, with categories designated as A, B, C, M, and X \cite{ahmadzadeh2021train}. Major flare events are intense solar phenomena, classified as M and X categories, that produce bursts of electromagnetic radiation across various wavelengths, from radio waves to gamma rays, and can release energy equivalent to billions of hydrogen bombs. Major flares can accelerate charged particles, generate solar energetic particle (SEP) events, and produce coronal mass ejections (CMEs) that travel through the solar system. When directed toward Earth, they can disrupt satellite operations, GPS and communication systems, lead to radiation risks for astronauts, and induce geomagnetic storms that affect power grids, posing significant threats to human endeavors and technological infrastructure \cite{hamdi2017time, bobra2015solar}. Today, a concrete theoretical link between the influx of magnetic fields and the onset of extreme solar events remains elusive. Therefore, considering the task as a data science challenge, focusing on the analysis of solar active region magnetic field parameters and utilizing machine learning techniques, is highly important in space weather research for improving predictions of solar flare events \cite{ahmadzadeh2021train}.

Research on machine learning-driven solar flare prediction has explored extensive methodologies, with recent advancements highlighting multivariate time series (MVTS) approaches that record photospheric magnetic field data from solar active regions over time. Recent models, built upon MVTS data, have demonstrated enhanced efficacy in predicting solar flaring activities compared to earlier models reliant on a single timestamp for magnetic field vector classification \cite{angryk2020multivariate}. For binary solar flare classification tasks, where major flare events are classified as flare positives and others as flare negatives, the infrequent occurrence of major flares leads to a pronounced class imbalance between flare-positive and flare-negative instances. This imbalance often leads to a tendency for models to overfit, as they may become biased towards the majority class. To address this challenge, a primary approach is to derive embeddings as a parametric mapping from the raw time series data to a feature vector that retains crucial characteristics. These new representations aim to improve transferability, enabling predictive models to perform better by capturing more abstract and useful patterns inherent in the data more effectively. Representing time series data as a vector of the latest timestamp \cite{bobra2015solar}, as a vector of descriptive statistics \cite{ahmadzadeh2021train}, or as a sequence and functional network embedding \cite{hamdi2022multivariate} are just a few examples of approaches to obtain different data representations. However, the use of contrastive learning methods in the time
series domain, particularly in solar flare tasks, remains relatively uncommon and is still in its early stages \cite{vural2024contrastive}. Most general contrastive learning methods focus on two key objectives: maximizing inter-class separation, where the embeddings of different classes are pushed further apart, and enhancing intra-class compactness, where embeddings of the same class are pulled closer together. This process encourages the model to learn discriminative representations, reduces the impact of majority class bias, and enables more effective performance in subsequent tasks \cite{le2020contrastive}. Guided by this approach, we aim to explore leveraging the foundational principles of contrastive learning to enhance representation learning for time series data, specifically targeting the issue of class imbalance in solar flare prediction.

In this paper, we present EXCON, a novel contrastive representation learning framework that leverages the key differences between distinct classes to tackle the challenges posed by severe class imbalance in time series data and significantly reduce overfitting to the majority class. In our methodology, we first obtain the dynamical features of each MVTS instance, representing them as vectors. Subsequently, a single \textit{extreme instance} is designated for each class, selected as the most distinctive representation within the feature space, thereby maximizing the separation between distinct class instances. Following this, we employ a temporal feature embedding module that generates refined embeddings, which are compelled to converge towards their corresponding extreme instances through the application of a custom extreme reconstruction loss function. These embeddings are subsequently set for deployment in downstream classification tasks, enabling more effective and discriminative performance. This unified approach offers a new perspective for handling solar flare prediction and facilitates robust representation learning across time series analysis tasks. The primary contributions of this work are outlined as follows:

\begin{itemize}[leftmargin=*]
    \item Proposing a novel contrastive learning approach to address the severe class imbalance. EXCON is versatile, applicable to univariate and multivariate time series data, and is effective for binary as well as multiclass classification problems.
    \item Designing a custom extreme reconstruction loss for the temporal feature embedding module, driving output embeddings towards their corresponding class extremes, thereby enhancing intra-class similarity.
    \item Measuring the performance of the proposed approach through metric analysis during the experimental phase, with a focus on metrics that align with the properties of the benchmark dataset. Additionally, the analysis includes evaluations using other archive datasets beyond the benchmark to ensure a comprehensive assessment.  
\end{itemize}

\section{Related Works} \label{sec:Related Work}

\subsection{Solar Flare Prediction}

Theophrastus (THEO) was among the initial efforts in predicting solar flares, an expert system based on sunspot classification. In 1987, The Space Environment Center (SEC) of the National Oceanic and Atmospheric Administration (NOAA) started the use of THEO officially \cite{mcintosh1990classification}. Following this, as multiple ground and space-based observatories collected vast amounts of magnetic field data, solar flare prediction evolved into a data-driven challenge. This shift led to the development of models for different types of data: models analyzing line-of-sight photospheric magnetic field parameters, and models examining the full-disk photospheric magnetic field to characterize active region parameters \cite{boubrahimi2016spatio}.

For over a decade, NASA's Solar Dynamics Observatory (SDO) has been continuously mapping the full-disk vector magnetic field at 12-minute intervals using the Helioseismic and Magnetic Imager (HMI) instrument \cite{mason2010testing}. Consequently, recent studies heavily rely on nonlinear statistical approaches, particularly predictive models, which frame solar flare prediction as a downstream classification task, utilizing continuous vector magnetogram data from SDO. Support vector machine \cite{bobra2015solar}, logistic regression \cite{song2009statistical}, decision tree \cite{yu2009short}, and fully connected neural network \cite{ahmed2013solar} are such examples. The study in \cite{angryk2020multivariate} extended single-timestamp models by introducing temporal window-based flare prediction. They developed Space Weather Analytics for Solar Flares (SWAN-SF), an MVTS dataset with magnetic field data recorded over a preset observation time, labeled with flare classes occurring after a specific prediction time. Since the introduction of SWAN-SF, diverse MVTS-based classification methodologies have developed. Extracting the latest timestamp of MVTS instances \cite{bobra2015solar}, MVTS decision trees with clustering as a preprocessing step \cite{ma2017solar}, LSTM-based deep sequence modeling for end-to-end flare classification with automated feature learning \cite{muzaheed2021sequence}, functional network embedding and sequence modeling to capture both temporal and spatial relationships of the MVTS instances \cite{hamdi2022multivariate} are examples of such methodologies. 

\subsection{Contrastive Representation Learning for Time Series}

In contrastive methods, learning of representation embeddings is achieved by enforcing that semantically similar samples are pulled closer, and embeddings of dissimilar samples are pushed apart \cite{le2020contrastive}. While contrastive learning methods have been extensively explored in applications of vision, language, and graphs, their integration into the time series domain remains relatively underexplored. Nevertheless, recent years have seen an increasing emphasis on applying contrastive learning techniques to time series data for the effective learning of representations. Accordingly, for time series, inter-sample relations were learned by sampling positive and negative samples from a given anchor, while intra-temporal relations were captured by sampling time pieces from the anchor to learn the underlying representations \cite{fan2020self}. Temporal and contextual contrastive learning methods enhanced time series representations by generating two views for each sample through strong and weak augmentations, incorporating both temporal and contrastive modules \cite{eldele2021time}. Criteria of high fidelity and variety were used to guide data augmentation selection, with a meta-learner automatically identifying suitable augmentations for different time series datasets \cite{luo2023time}. A temporal and instance-wise contrastive loss with soft assignments was designed to prevent degradation of learned representations from contrasting similar instances or adjacent timestamps \cite{lee2023soft}. A Siamese architecture with a convolutional encoder was employed for time series forecasting without needing negative pairs \cite{zheng2023simts}. These recent developments signal the growing adoption of contrastive learning in time series representation, suggesting its expanding influence and potential.

\section{Dataset} \label{sec:Dataset}

The SWAN-SF dataset, introduced in 2020, has become a significant asset for advancing solar flare research, driving time series-based prediction methods forward. Compiled from vector magnetograms of the Sun’s photosphere in the SHARP (Space-weather HMI Active Region Patch) series, SWAN-SF provides a substantial archive of MVTS data from active regions observed between 2010 and 2018 \cite{angryk2020multivariate}. MVTS instances within SWAN-SF are labeled with one of five distinct categories of flare events: FQ (combining flare quiet and A category events), B, C, M, and X, with intensities increasing respectively. In solar flare prediction tasks, it is a common approach to group the mentioned categories under two classes: non-flare (NF) and flare (F). The NF class includes smaller flare events, such as FQ, B, and C categories, while the F class covers major flare events, namely M and X categories, which pose threats to public health and can cause severe disruptions to infrastructure \cite{hamdi2017time, bobra2015solar}.

Each data instance in the SWAN-SF dataset represents an MVTS slice containing 24 photospheric magnetic field parameters from solar active regions, extracted using a sliding window approach \cite{bobra2015solar}. For a given flare with a unique ID, multiple equal-length MVTS slices are extracted, each covering a time frame called the observation window ($T_{\text{obs}}$). The starting point of each subsequent slice is determined by $s_i + \tau$, where $s_i$ is the start of the $i^{\text{th}}$ segment and $\tau$ is the step size. Each slice is labeled based on the most intense flare observed within a predefined prediction window ($T_{\text{pred}}$) that follows $T_{\text{obs}}$. In SWAN-SF, $T_{\text{obs}}$ and $T_{\text{pred}}$ span 12 and 24 hours, respectively, with $\tau$ set to 1 hour \cite{ahmadzadeh2021train, angryk2020multivariate}. Each instance, denoted as $mvts^{\text{(m)}} \in \mathbb{R}^{\tau \times N}$, represents a univariate time series of length $\tau$ for each of the $N$ magnetic field parameters, where $1 \leq m \leq M$ and $M$ is the total number of instances.

\begin{figure}
\centering
\includegraphics[width = 0.92\linewidth]{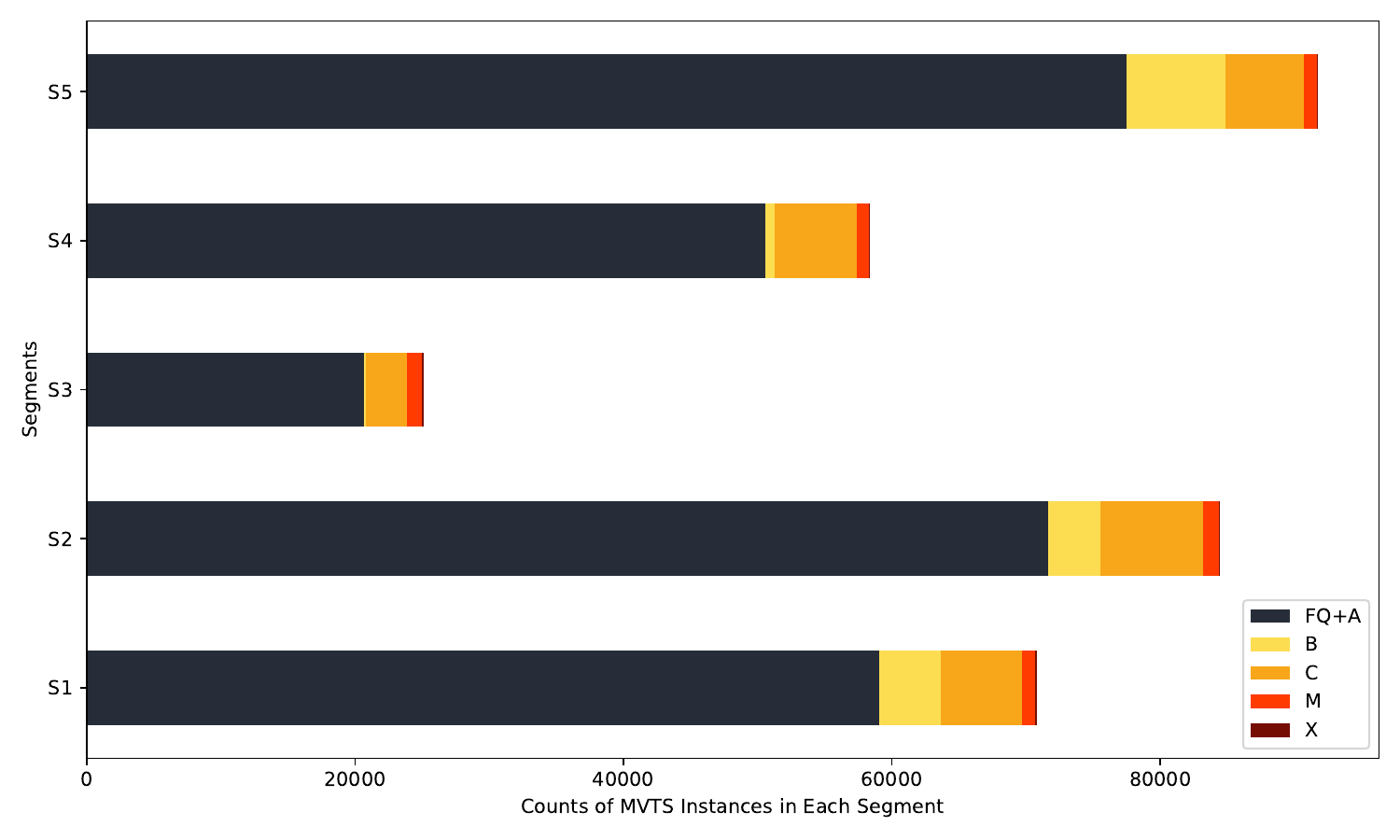}
\caption{Segments of SWAN-SF benchmark dataset with the frequencies of five flare categories indicated in each segment.}
\label{fig:mvts_partitions}
\end{figure}

SWAN-SF includes multiple segments covering different observation periods (i.e., S1, S2, ..., S5). Fig. \ref{fig:mvts_partitions} offers insight into the class distribution. In SWAN-SF, there is a severe imbalance between NF and F classes as a result of the infrequent occurrence of major flare events. This disparity between NF and F examples often leads to classification results biased towards the majority class, thereby yielding high true negative rates and low true positive rates. For precise and objective classification evaluation, it is essential to select appropriate metrics, as discussed in Section \ref{sec:Experimental Evaluation}.

\section{Methodology} \label{sec:Methodology}

\subsection{Extraction of Dynamical Features}
\begin{figure}
\centering
\includegraphics[width = 0.75\linewidth]{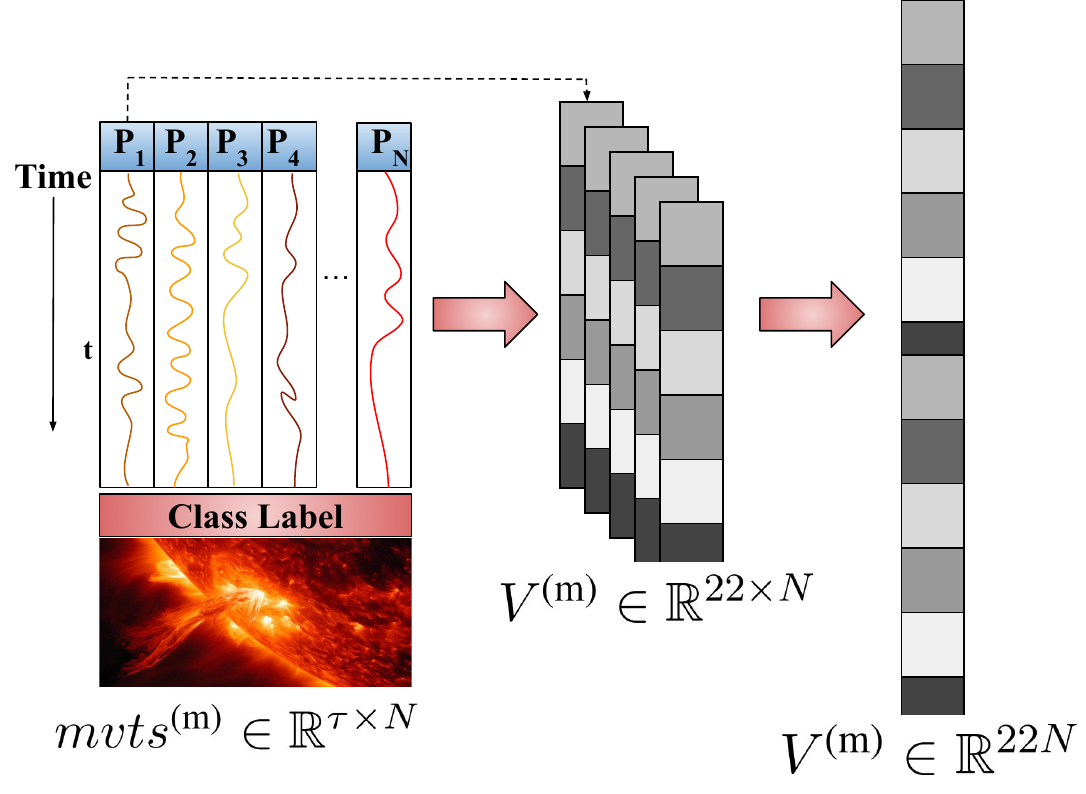}
\caption{MVTS feature extraction process with catch22.}
\label{fig:c22_extraction}
\end{figure}

In our approach, the initial step is to compress MVTS data instances into a vector representation. To achieve this, we utilize the catch22 feature extraction method \cite{lubba2019catch22} that condenses time series data into 22 distinct features, providing a concise and interpretable representation of the dynamical characteristics of each MVTS instance. These features encompass the distribution of values in the time series, temporal autocorrelation properties, scaling of fluctuations, and others. The catch22 method has proven effective in generating feature representations that are both highly discriminative and minimally redundant, showing strong performance across University of East Anglia (UEA) MVTS classification archive datasets \cite{ruiz2021great, bagnall2018uea}. We apply catch22 feature extraction to each univariate time series within the MVTS instances as demonstrated in Fig. \ref{fig:c22_extraction}. Consequently, for each data instance $mvts^{\text{(m)}} \in \mathbb{R}^{\tau \times N}$, where $N$ is the number of parameters and $\tau$ is the time series length, we extract a compressed multi-catch22 vector $V^{\text{(m)}} \in \mathbb{R}^{d}$. Hence, $d$, the length of the multi-catch22 vector, is equal to $22N$ \cite{vural2024contrastive}.

\subsection{Derivation of Extreme Instances}
\begin{figure}
\centering
\includegraphics[width = 0.75\linewidth]{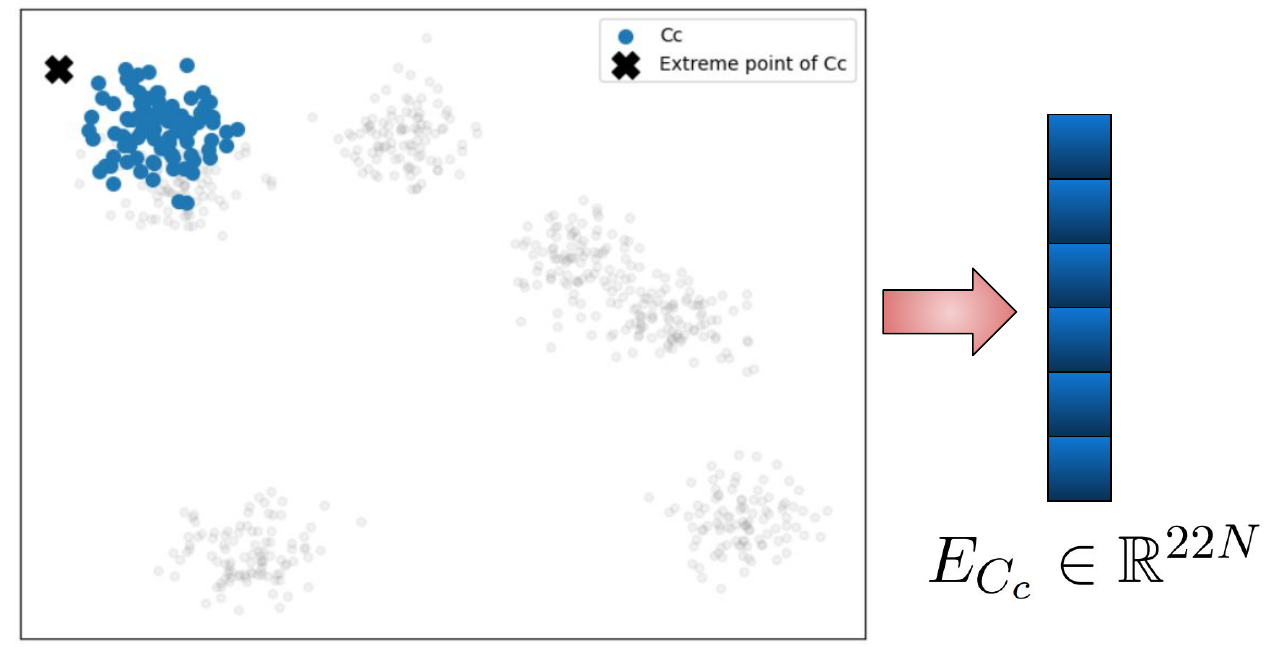}
\caption{Process of deriving extreme instance of class $C_c$.}
\label{fig:Extreme_point}
\end{figure}

Having acquired the multi-catch22 vectors $V^{\text{(m)}} \in \mathbb{R}^{d}$, the next step in our approach involves identifying extreme instances to serve as distinctive representations for each class. This step is crucial for enhancing the contrastive abilities of the model by effectively drawing data instances closer to their corresponding extreme representations. To achieve this, we assign an extreme instance to each class $C_c$ by determining the multi-catch22 vector that maximizes the complete linkage, thereby identifying the data instance that exhibits the greatest distance from the other class clusters. This approach adopts a one-versus-all strategy, wherein the extreme representation $E_{C_c}$ for class $C_c$ is selected for its ability to enhance the inter-class separation in the feature space as Fig. \ref{fig:Extreme_point} shows.

\begin{equation}
D(C_c, N) = \max_{V_{C_c} \in C_c, V_{N} \in N} d(V_{C_c}, V_{N})  \label{eq:extremes}
\end{equation}

In Equation \ref{eq:extremes}, \( D(C_c, N) \) is the distance between the cluster of class \( C_c \) and the cluster of all other classes \( N \), and \( d(V_{C_c}, V_{N}) \) is the Euclidean distance between the multi-catch22 vector \( V_{C_c} \in \mathbb{R}^{d} \) of class \( C_c \) and the multi-catch22 vector \( V_{N} \in \mathbb{R}^{d} \) of the set of all other classes \( N \).  

\subsection{Framework}
\begin{figure*}[t]
\centering
\includegraphics[width=0.85\linewidth]{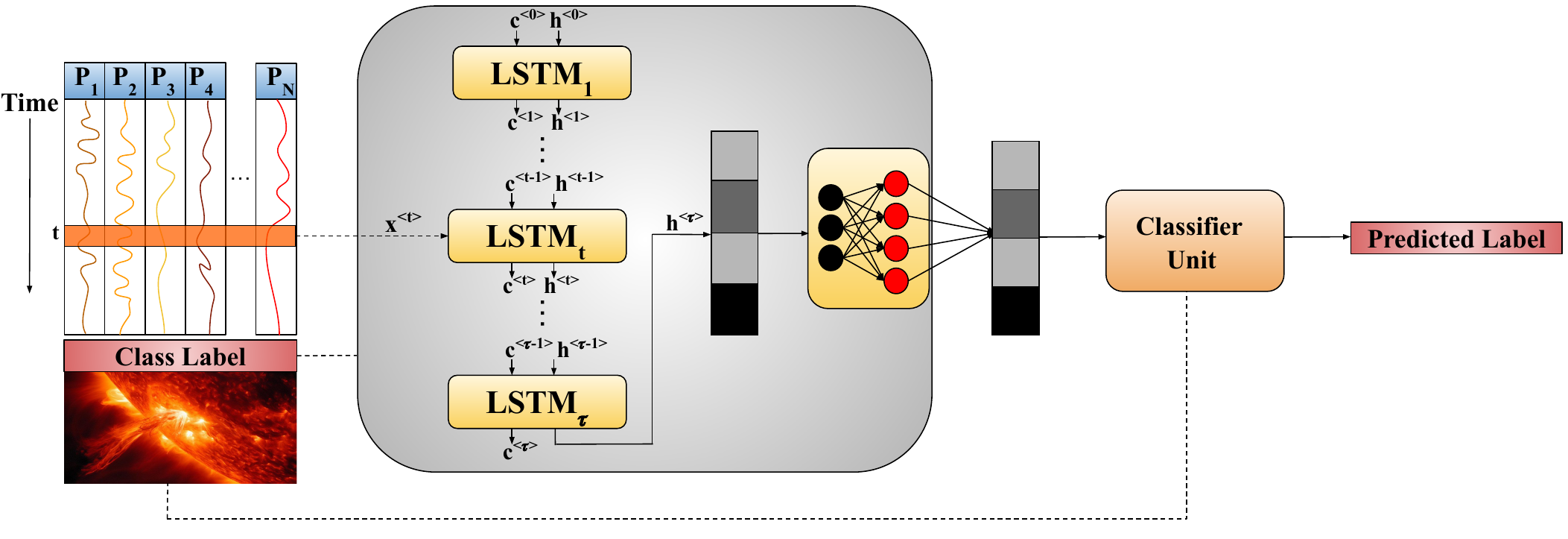}
\caption{In the EXCON framework, at timestamp $t$, vector $x^{<t>}$ of MVTS instance is processed by the $t^{th}$ LSTM cell within temporal feature embedding module. In the last timestamp $\tau$, the output $h^{<\tau>}$ is projected into \(d\)-dimensional space by the fully connected layer. The downstream classifier utilizes the learned embeddings for class prediction.}
\label{fig:framework}
\end{figure*}

EXCON comprises two integrated phases that function in an end-to-end framework. The first phase learns embeddings from MVTS data instances. The second phase utilizes these embeddings to perform the classification task. For a detailed illustration of our framework, please refer to Fig. \ref{fig:framework}.

\begin{itemize}[leftmargin=*]
\item \noindent \textbf {Temporal Feature Embedding Module:}

In the temporal feature embedding module, the processing of MVTS data instances is carried out using a long short-term memory (LSTM) layer. Each instance, denoted as $mvts^{\text{(m)}} \in \mathbb{R}^{\tau \times N}$, is interpreted as a sequence comprising $\tau$ timestamp vectors, where each vector $x^{<t>} \in \R^{N}$ represents the data at a specific time step. These timestamp vectors are sequentially processed through the LSTM cells, which are adept at capturing temporal dependencies within the data. The input to the LSTM layer corresponds to $N$ parameters, which define the dimensionality of the time series at each time step. The final hidden state representation $h^{<\tau>}$ of the LSTM, after processing all timestamp vectors, is used to produce an internal embedding vector. This embedding vector is then projected into a \(d\)-dimensional space, matching the size of the multi-catch22 vectors. This projection is achieved through a single fully connected neural network layer. To introduce regularization to the network and prevent overfitting, a single dropout layer is incorporated. Upon completion of the training process, the embedding vector \( e_{label}^{\text{(m)}} \in \mathbb{R}^{d} \) for the \(m^{\text{th}}\) MVTS instance can be extracted from the final layer of the network. This vector represents the learned features of the MVTS instance in the reduced \(d\)-dimensional space, which is crucial for subsequent tasks such as classification or further analysis.

\item \noindent \textbf {Extreme Reconstruction Loss:}

Here, we introduce our novel extreme reconstruction loss function to train our temporal feature embedding module. This loss function is designed to enforce that the embeddings learned by the model are aligned with the extremes of each class in a supervised setting, inspired by \cite{vural2024contrastive} and extended to handle the multiclass scenario. Specifically, for each embedding vector $e_{C_c}^{(m)} \in \mathbb{R}^{d}$ associated with a class $C_c$, we compute the mean squared error (MSE) loss relative to the corresponding class extreme $E_{C_c} \in \mathbb{R}^{d}$. The purpose of this approach is to facilitate effective contrastive learning for MVTS data instances. By doing so, it ensures that MVTS instances belonging to the same class are embedded closer to their respective class extremes in the new feature space, thereby minimizing intra-class variability. 
Conversely, due to the influence of extreme points, data instances from differing classes are increasingly separated in the embedding space, enhancing inter-class distinctiveness. This personalized loss function thus helps in fine-tuning the embeddings to capture the distinctive characteristics between different class instances. By guiding the embeddings to be closely aligned with class extremes and ensuring proper separation between classes, we enhance the discriminative power of the learned representations, ultimately facilitating more accurate classification. The extreme reconstruction loss, therefore, plays a crucial role in improving the model's performance by emphasizing the key differences between classes in the representation space.

\begin{equation}
\begin{aligned}
\mathcal{L}_{\text{ER}} = \sum_{c=1}^{C} \frac{1}{|C_c|} \sum_{m=1}^{|C_c|} \sum_{i=1}^{d} \left(e_{C_c}^{(m)}[i] - E_{C_c}[i]\right)^2
\end{aligned}
\label{eq:extreme_reconstruction_loss}
\end{equation}

In Equation \ref{eq:extreme_reconstruction_loss}, $C$ is the number of classes, $|C_c|$ is the number of instances belonging to class $C_c$, $d$ is the dimension of embedding vectors, $e_{C_c}^{(m)}[i]$ is the $i^{th}$ entry of $m^{th}$ embedding vector, $E_{C_c}[i]$ is the $i^{th}$ entry of $C_c$ class extreme.

\item \noindent \textbf {Classification of Learned Embeddings:}
In the second phase of our framework, the embedding vectors $e_{C_c}^{\text{(m)}} \in \mathbb{R}^{d}$, which have been extracted, are employed as input data to train a subsequent classifier unit in a supervised manner to generate the final class predictions. The decision regarding which downstream classifier to use is intricately tied to the nature of the classification problem at hand and is elaborated upon in Section \ref{sec:Experimental Evaluation}.
\end{itemize}

\section{Experimental Evaluation} \label{sec:Experimental Evaluation}
Here, we evaluate the EXCON framework based on its classification performance on both the SWAN-SF dataset and multiple time series archive datasets. The source code for all experimentation phases is available on our GitHub repository\footnote{\url{https://github.com/OnurVural/excon}}.

\subsection{Evaluation Metrics}  \label{sec:eval}
Given the severe class imbalance between NF and F data instances in the SWAN-SF dataset, relying solely on accuracy—focusing only on correct predictions—is insufficient. Consequently, we employ several supplementary metrics commonly used in current literature to assess the flare prediction performance. True Skill Statistic (TSS) effectively addresses class imbalance by quantifying the difference between true positive and false positive rates, with values ranging from -1 to 1. TSS is therefore recommended as the main metric for evaluating flare prediction performance \cite{bobra2015solar}. Heidke Skill Score (HSS2) measures the model's improvement over random predictions. The F1 score assesses the model's ability to correctly classify positive instances by balancing precision and recall. Gilbert Skill Score (GS) estimates the probability of obtaining true positives by chance. Receiver Operating Characteristic Area Under the Curve (ROC AUC) evaluates the classifier's capacity to differentiate between classes at various threshold settings \cite{hamdi2017time, bobra2015solar, mason2010testing}. 

\subsection{Preprocessing of MVTS instances}

We included only the FQ category MVTS instances from the NF class in the training sets, as the B and C categories exhibit significant magnetic field parameter similarities with the M and X categories due to shared solar activity patterns \cite{ahmadzadeh2021train}. We applied fast Pearson correlation-based k-nearest neighbors (FPCKNN) imputation \cite{eskandarinasab2024impacts} to address any missing values. Following this, we performed instance-wise normalization of the MVTS instances across the individual time series features. Given an instance \( mvts^{\text{(m)}} \in \mathbb{R}^{\tau \times N} \), having $\tau$ timestamps and $N$ parameters corresponding to individual time series (i.e., $P_{1}$, $P_{2}$, ... $P_{N}$), the normalization is executed for each parameter across the entire time series.

\begin{equation}
x_{n}^{<t>} = \frac{x_{n}^{<t>} - \mu_{n}}{\sigma_{n}} \label{eq:znorm}
\end{equation}

In equation \ref{eq:znorm}, $x_{n}^{<t>}$ is the $t^{th}$ timestamp entry of the univariate time series $P_{n}$, where $1 < t < \tau$, $\mu_{n}$ and $\sigma_{n}$ are the mean and standard deviation of time series $P_{n}$ respectively.

\subsection{Component Selection and Training of EXCON}

\begin{figure}
\centering
\includegraphics[width=0.99\linewidth]{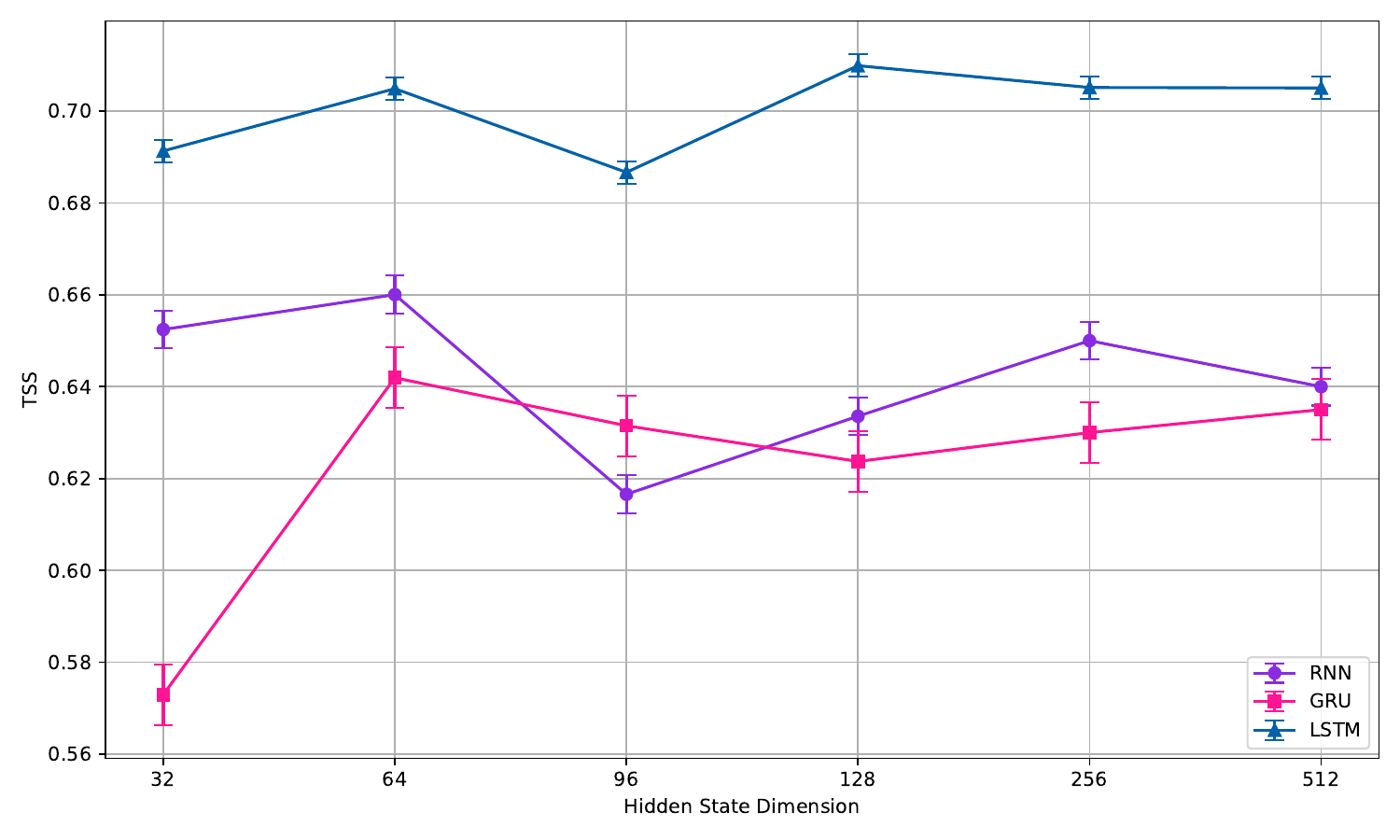}
\caption{Comparison of mean TSS performance for selected sequence models within the temporal feature embedding module of the EXCON framework across different hidden dimensions.}
\label{fig:roc_hidden}
\end{figure}

We selected successive SWAN-SF segments in the training and testing stages (e.g. S1 to derive extremes and training, S2 for evaluation). We performed experiments to assess the impact of the main components within the EXCON framework. Initially, we evaluated logistic regression (LR), support vector machine (SVM), k-nearest neighbors (KNN) with k = 5, decision tree (DT), and multilayer perceptron (MLP) with one hidden layer size 100 as the classifier unit. LR emerged as the top-performing classifier, achieving the best mean performance in three consecutive train-test segments, as detailed in Table \ref{table:downstream}. After choosing the classifier unit, we investigated the sequence modeling of MVTS data instances in the temporal feature embedding module by comparing the performance of LSTM-based modeling against Recurrent Neural Network (RNN) and Gated Recurrent Unit (GRU). This comparison also included experimenting with different dimensions of hidden spaces to determine their effects on model performance. As shown in Fig. \ref{fig:roc_hidden}, the LSTM model achieved the best mean performance across segments S1-S2, S2-S3 and, S3-S4. In its final configuration, the EXCON framework is set with the following hyperparameters: LSTM cells with an input dimension of 24 (representing number of parameters in MVTS), a hidden state dimension of 128 (chosen from options 32, 64, 128, 256, and 512), a dropout rate of 0.5, a single fully connected layer (selected from one, two, or three-layer configurations) with an input size of 128 and an output size of 528 (d-dimension), the Adam optimizer with a learning rate of $10^{-2}$ (selected from $10^{-1}$, $10^{-2}$, $10^{-3}$, and $10^{-4}$), and 30 training epochs.

\begin{table*}
\centering
\caption{Solar flare prediction performance of selected classifier units in EXCON}
\label{table:downstream}
\small
\begin{tabular}{lccccccc}
\toprule
Classifier & Accuracy & TSS & HSS2 & F1 & GS & ROC AUC \\
\midrule
LR & \textcolor{red}{\textbf{0.7306 ± 0.09662}} & \textcolor{red}{\textbf{0.7098 ± 0.09779}} & \textcolor{red}{\textbf{0.1189 ± 0.04965}} & \textcolor{red}{\textbf{0.1579 ± 0.04766}} & 0.02303 ± 0.007004 & \textcolor{red}{\textbf{0.8549 ± 0.04889}}\\
SVM & 0.6819 ± 0.1344 & 0.6645 ± 0.1362 & 0.1026 ± 0.055  & 0.1428 ± 0.051 & 0.02318 ± 0.007065
& 0.8322 ± 0.06812\\
KNN & 0.7163 ± 0.125 & 0.693 ± 0.1248 & 0.1182 ± 0.06089 & 0.1574 ± 0.0564 & 0.02297 ± 0.007
& 0.8465 ± 0.06242\\
DT	& 0.7216 ± 0.1354 & 0.6318 ± 0.1176 & 0.1113 ± 0.05374 & 0.1507 ± 0.04774 & 0.02068 ± 0.006054 & 0.8159 ± 0.05879\\
MLP & 0.6805 ± 0.08599 & 0.6671 ± 0.09098 & 0.09438 ± 0.03769 & 0.1348 ± 0.03946 & \textcolor{red}{\textbf{0.02322 ± 0.006865}} & 0.8336 ± 0.04549\\
\bottomrule
\end{tabular}
\end{table*}

\subsection{Analysis of Learned Embeddings}
\begin{figure}
\centering
\includegraphics[width=0.99\linewidth]{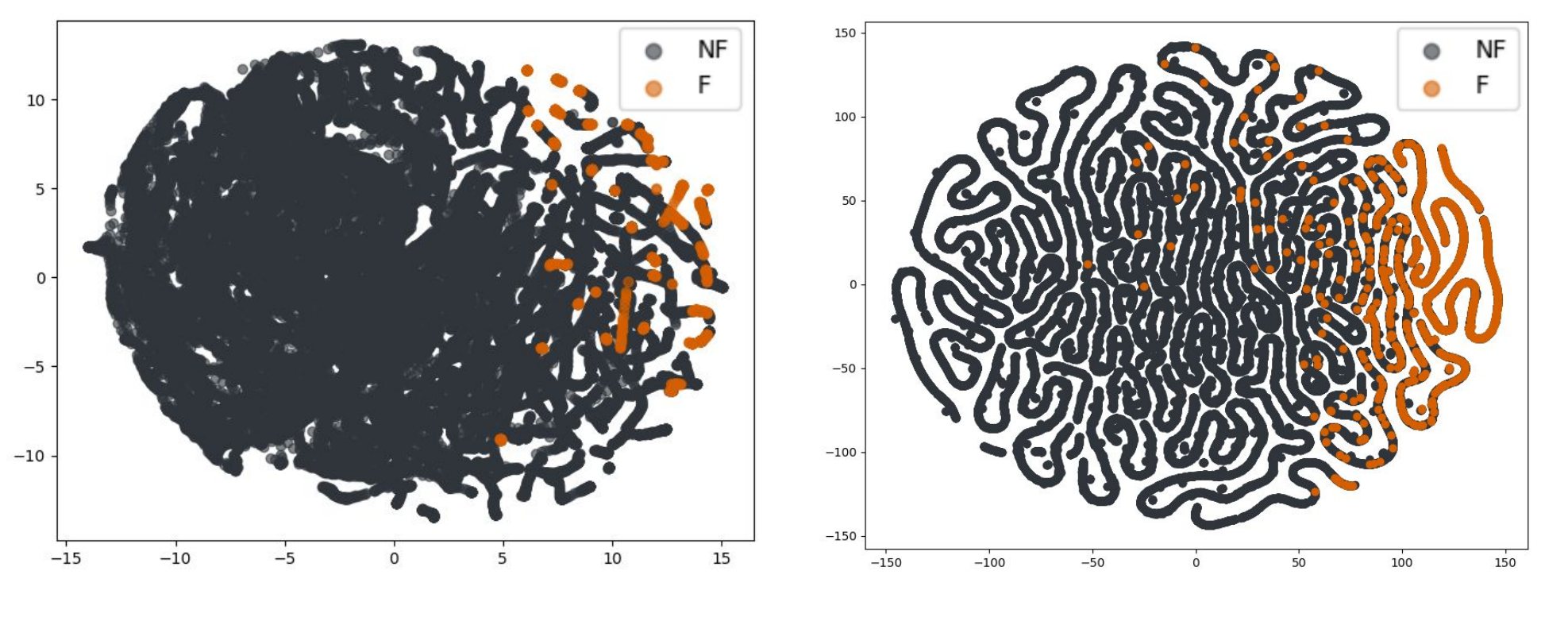}
\caption{t-SNE visualization of Segment 2 as test data: raw format and after obtaining the embeddings}
\label{fig:tsne_centroid}
\end{figure}

To visually analyze how well the learned embeddings distinguish between the classes, we employ t-SNE visualization \cite{van2008visualizing} to project the extracted embeddings, denoted as \( e_{label}^{\text{(m)}} \in \mathbb{R}^{d} \), into a two-dimensional space. In Fig. \ref{fig:tsne_centroid}, compared to the raw SWAN-SF data distribution, the new feature space created by our embeddings shows highly improved separation between minor and major flare events, as demonstrated by the distinct clustering of these classes. This result highlights EXCON’s ability to capture relevant features that differentiate between these classes, thus demonstrating the robustness of our approach in learning meaningful representations.

\subsection{Tracing Input}
To trace the data flow in our framework, we use a single instance of our MVTS dataset. Specifically, we examine an MVTS instance corresponding to a time slice that captures the observation of a significant X-category solar flare, famously known as the Valentine's Day Flare, which erupted on February 14, 2011. This particular instance, denoted as $mvts_{F}^{\text{(m)}} \in \mathbb{R}^{60 \times 24}$ is processed through our temporal feature embedding module to transform into the new embedding $e_{F}^{(m)} \in \mathbb{R}^{528}$. This new embedding is subsequently fed into our downstream classification unit. The model successfully classifies this embedding, accurately identifying the Valentine's Day Flare as a major flare of the F class. This successful classification demonstrates the effectiveness of our framework in capturing and interpreting complex solar flare events.

\subsection{Baselines}
To assess the effectiveness of EXCON, we compare it against several established approaches from the existing literature on solar flares, each of which considers MVTS instances as a different data representation. For consistency, we use LR as the downstream classifier when required.

\begin{itemize} \item \noindent \textbf{MVTS to Vector (MVTS2V)}: We flatten each instance $mvts^{\text{(m)}} \in \mathbb{R}^{\tau \times N}$ into a vector $vmvts^{\text{(m)}} \in \mathbb{R}^{\tau N}$ and use it to train a downstream classifier \cite{bobra2015solar}. 

\item \noindent \textbf{Latest Parameter Value Vector (LPVV)}: We collect the magnetic field parameters from the latest timestamp of the MVTS, denoted as $x^{<t>} \in \mathbb{R}^{N}$, which is closest in time to the flaring event, to train a downstream classifier \cite{bobra2015solar}. 

\item \noindent \textbf{Sequence Modeling (SEQ)}: We feed MVTS instances only into a sequence model (i.e., LSTM) as timestamp vectors at each time step, $x^{<t>} \in \mathbb{R}^{N}$, and use the last hidden state as the final representation \cite{muzaheed2021sequence}. We use a hidden state dimension of 128 and a learning rate of $10^{-2}$. 

\item \noindent \textbf{Random Convolutional Kernel Transform (ROCKET)}: We use a dedicated time series classifier, ROCKET, that utilizes randomly generated convolutional kernels for each $mvts^{\text{(m)}} \in \mathbb{R}^{\tau \times N}$. It has demonstrated high performance on UEA MVTS datasets \cite{dempster2020rocket, ruiz2021great}. 

\end{itemize}

\subsection{Classification Performance on SWAN-SF Dataset}

\begin{figure}
\centering
\includegraphics[width=0.99\linewidth]{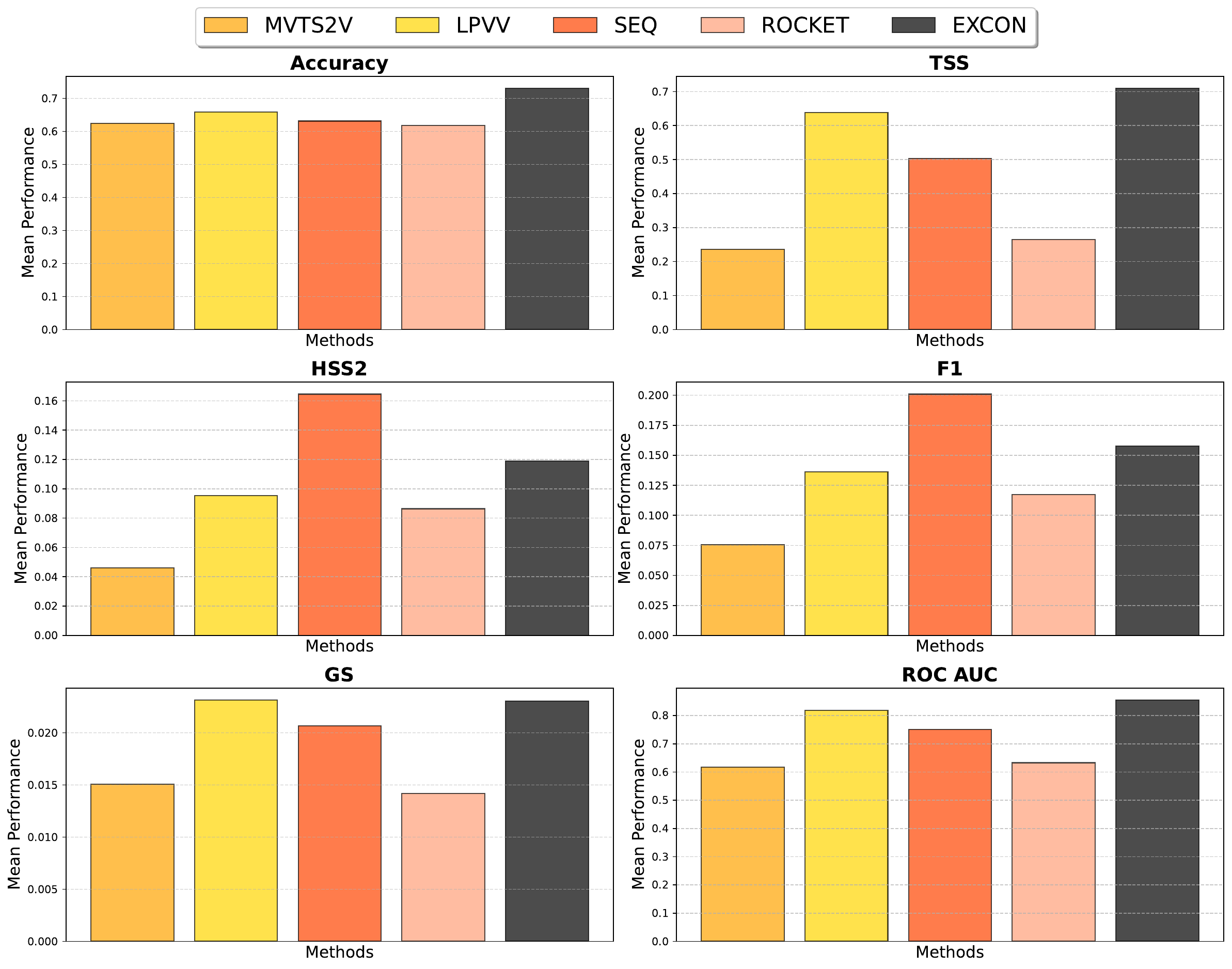}
\caption{Bar chart of EXCON vs. baselines in SWAN-SF solar flare prediction performance. }
\label{fig:performance}
\end{figure}

After our extensive experimentation phase using three consecutive segments as train-test pairs (i.e., S1-S2, S2-S3, and S3-S4), the mean performance scores are summarized in Table \ref{table:comparison} and illustrated in Fig. \ref{fig:performance}. These scores reveal that EXCON exhibits notable contrastive capabilities, achieving strong performance results. EXCON stands out as the best performer with an accuracy of 0.7306, a TSS of 0.7098, and a ROC AUC of 0.8549. This performance surpasses that of LPVV by 5.1\%, 7.2\%, and 3.6\%, respectively. However, when it comes to HSS2 and F1 score, EXCON is the second best performer, trailing behind the leading performer, SEQ, by 4.6\% and 4.3\%, respectively. In terms of GS, EXCON ties with LPVV, with a very slight margin of 0.011\% between them. These results highlight EXCON's strong competitive edge against state-of-the-art methods, showcasing its effectiveness while also identifying areas for potential improvement.

\begin{table*}
\centering
\caption{Performance comparison of baseline methods for solar flare prediction}
\label{table:comparison}
\small
\begin{tabular}{lccccccc}
\toprule
Method & Accuracy & TSS & HSS2 & F1 & GS & ROC AUC \\
\midrule
EXCON & \textcolor{red}{\textbf{0.7306 ± 0.09662}} & \textcolor{red}{\textbf{0.7098 ± 0.09779}} & 0.1189 ± 0.04965 & 0.1579 ± 0.04766 & 0.02303 ± 0.007004 & \textcolor{red}{\textbf{0.8549 ± 0.04889}}\\
MVTS2V & 0.6243 ± 0.4204 &	0.2358 ± 0.3335 & 0.04592 ± 0.06494 & 0.07575 ± 0.06675	& 0.01507 ± 0.01333 & 0.6179 ± 0.1667\\
LPVV & 0.6589 ± 0.1728 & 0.6378 ± 0.1748 & 0.09523 ± 0.05173 & 0.1361 ± 0.04493 
& \textcolor{red}{\textbf{0.02314 ± 0.007139}} & 0.8189 ± 0.08738\\
SEQ & 0.6314 ± 0.2914 & 0.5027 ± 0.1967 & \textcolor{red}{\textbf{0.1646 ± 0.1766}} & \textcolor{red}{\textbf{0.2009 ± 0.1626}} & 0.02068 ± 0.008318 & 0.7514 ± 0.09834 \\
ROCKET & 0.6174 ± 0.2688 & 0.2652 ± 0.162 & 0.08639 ± 0.07387 & 0.1174 ± 0.06113 & 0.01416 ± 0.01185 & 0.6326 ± 0.08102  \\
\bottomrule
\end{tabular}
\end{table*}

\subsection{Classification Performance on UEA Benchmark Datasets}

After completing experiments with the SWAN-SF benchmark dataset, we expand our analysis to evaluate the performance of EXCON across selected UEA archive benchmark datasets \cite{bagnall2018uea}. While choosing the datasets, we aimed to ensure a broader representation of different problem domains and data types. To achieve this, we carefully chose two datasets from each category: univariate time series for binary classification, multivariate time series for binary classification, univariate time series for multiclass classification, and multivariate time series for multiclass classification. Table \ref{table:uae_data} provides a detailed summary of each UAE dataset, highlighting key characteristics such as the size of the training and test sets, the time series length ($\tau$), and the number of features ($N$). Here, we exclude LPVV from consideration due to its primary applicability to the SWAN-SF dataset. Additionally, we report the results using the best-performing downstream classifier for each dataset.

\begin{figure}
\centering
\includegraphics[width=0.99\linewidth]{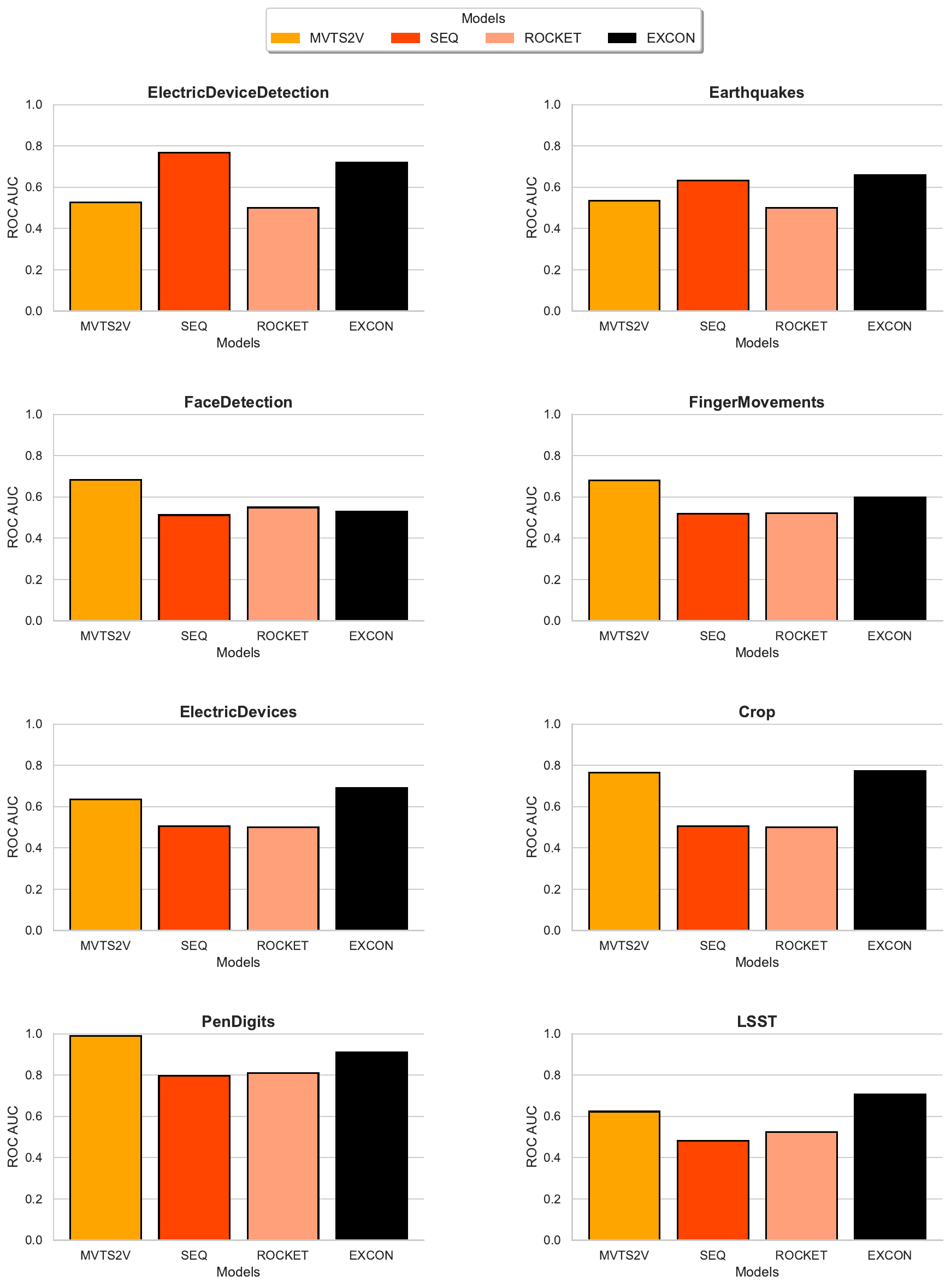}
\caption{Bar chart of EXCON vs. baselines in ROC AUC performance across UEA datasets. }
\label{fig:uea_roc_performance}
\end{figure}

Fig. \ref{fig:uea_roc_performance} shows one-versus-all ROC AUC results of our baselines across selected datasets. Through this experiment, several meaningful insights can be drawn. First, EXCON demonstrates competitive performance across a wide range of time series classification tasks and shows the best performance in four datasets, both binary and multiclass, indicating its robustness and versatility. However, it is worth noting that there are datasets where EXCON's performance lags behind other models. EXCON was initially designed to target the extreme differences between classes, as observed in the SWAN-SF benchmark dataset. Therefore, not all datasets may exhibit such pronounced distinctions, which could affect the performance. Additionally, the current experiment used data under relatively crude conditions without extensive preprocessing, which may impact the effectiveness of EXCON. Moving forward, we aim to address these limitations by incorporating advanced preprocessing techniques and refining the model to provide greater capabilities across diverse time series tasks.

\begin{table}
\centering
\caption{Selected UEA benchmark datasets}
\label{table:uae_data}
\small
\begin{tabular}{lcccccc}
\toprule
Dataset & Train & Test & $\tau$ & $N$ & Classes \\
\midrule
Earthquakes & 322 &	139 & 512 &	1 &	2\\
ElectricDeviceDetection & 623 & 3767 & 256 & 1 & 2\\
FaceDetection & 5890 &	3524 &	62 & 144 & 2\\
FingerMovements & 316 &	100 &	50 &	28	& 2\\ 
ElectricDevices & 8926 & 7711 & 96 & 1 & 7\\
Crop & 7200 & 16800 & 46 & 1 & 24\\ 
PenDigits & 7494 & 3498 &8 & 2 & 10\\ 
LSST & 2459 & 2466 & 36 & 6 & 14\\
\bottomrule
\end{tabular}
\end{table}

\section{Conclusion} \label{sec:Conclusion}
In this work, we presented a contrastive learning-focused method tailored for time series classification, with a particular focus on solar flare prediction tasks. Our methodology involved several steps: first, we captured relevant attributes of MVTS instances. Next, we computed contrastive extreme instances using these attributes to highlight meaningful distinctions between classes. As the core of our approach, we performed temporal embedding extraction for the MVTS data instances using our extreme reconstruction loss. This loss is designed to leverage the contrastive extreme instances to effectively separate each class, producing embeddings that lead to robust classification results in downstream tasks. Our framework demonstrated promising performance in time series classification experiments, including MVTS-based solar flare prediction and tasks on archive datasets.

Looking ahead, we are committed to advancing the capabilities of our framework and improving its performance across a broader range of time series classification tasks. This study aimed to address the problem in its original form, acknowledging class imbalance. Future work will explore the effectiveness of our framework with class imbalance mitigated by various undersampling and oversampling techniques. A key direction is incorporating additional concepts of contrastive learning into our custom loss function to focus specifically on inter-class separation. This includes adding a margin-based constraint for instances from different classes, penalizing different class embeddings that are too close to each other, applying orthogonality regularization, and introducing class weights based on the class distribution in the embedding space. While our current analysis has been limited to a subset of UEA benchmark datasets and performance metrics, we plan a comprehensive study that explores a wider range of commonly used benchmarks and employs rigorous evaluation methods.

\section*{Acknowledgment}
This project has been supported in part by funding from CISE and GEO Directorates under NSF awards \#2204363, \#2240022, \#2301397, and \#2305781.

\bibliographystyle{IEEEtran}
\bibliography{excon_contrastive_learning_time_series_BigData24.bib}

\end{document}